\documentclass[a4paper,fleqn]{cas-dc}

\usepackage{soul}
\usepackage{color}
\usepackage[numbers]{natbib}
\usepackage{subfigure}
\usepackage{graphicx}
\usepackage{caption}

\def\tsc#1{\csdef{#1}{\textsc{\lowercase{#1}}\xspace}}
\tsc{WGM}
\tsc{QE}
\tsc{EP}
\tsc{PMS}
\tsc{BEC}
\tsc{DE}

\begin{document}
\let\WriteBookmarks\relax
\def\floatpagepagefraction{1}
\def\textpagefraction{.001}
\shorttitle{Person Image Generation with Semantic Attention Network for Person Re-identification}
\shortauthors{Meichen Liu et~al.}


\title [mode = title]{Person Image Generation with Semantic Attention Network for Person Re-identification}                
\tnotemark[1]

\tnotetext[1]{This document is the results of the research
   project funded by the National Science Foundation.}


\author[1]{Meichen Liu}[type=editor,
                        auid=000,bioid=1,
                        orcid=0000-0002-3242-783X]
\ead{meichen@hrbeu.edu.cn}


\address[1]{College of Automation, Harbin Engineering University, Harbin 150001, China}

\author
[1]
{Kejun Wang}
\cormark[1]
\ead{heukejun@126.com}
\ead[URL]{http://homepage.hrbeu.edu.cn/web/wangkejun}

\author
[2]
{Juihang Ji}
\ead{jiruihang@hit.edu.cn}
\address[2]{College of Automation, Harbin Institute of Technology, Harbin 150001, China}

\author[3]{Shuzhi Sam Ge}[%
   ]
\ead{samge@nus.edu.sg}
\ead[URL]{https://robotics.nus.edu.sg/sge/}


\address[3]{National University of Singapore
, 117576, Singapore}

\cortext[cor1]{Corresponding author}


\begin{abstract}
Pose variation is one of the key factors which prevents the network from learning a robust person re-identification (Re-ID) model. To address this issue, we propose a novel person pose-guided image generation method, which is called the semantic attention network. The network consists of several semantic attention blocks, where each block attends to preserve and update the pose code and the clothing textures. The introduction of the binary segmentation mask and the semantic parsing is important for seamlessly stitching foreground and background in the pose-guided image generation. Compared with other methods, our network can characterize better body shape and keep clothing attributes, simultaneously. Our synthesized image can obtain better appearance and shape consistency related to the original image. Experimental results show that our approach is competitive with respect to both quantitative and qualitative results on Market-1501 and DeepFashion. Furthermore, we conduct extensive evaluations by using
person re-identification (Re-ID) systems trained with the pose-transferred person based augmented data. The experiment shows that our approach can significantly enhance the person Re-ID accuracy.

\end{abstract}

\begin{graphicalabstract}
\includegraphics{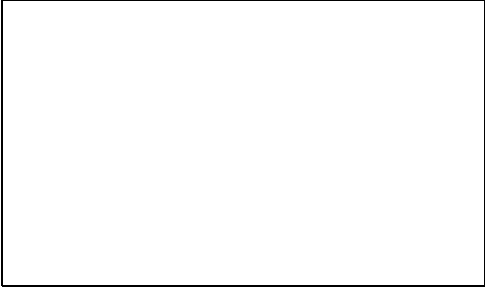}
\end{graphicalabstract}

\begin{highlights}
\item Research highlights item 1
\item Research highlights item 2
\item Research highlights item 3
\end{highlights}

\begin{keywords}
semantic parsing \sep pose transfer \sep image generation \sep person re-identification
\end{keywords}

\maketitle

\section{Introduction}
\begin{figure}
\centering
\includegraphics[scale=0.15]{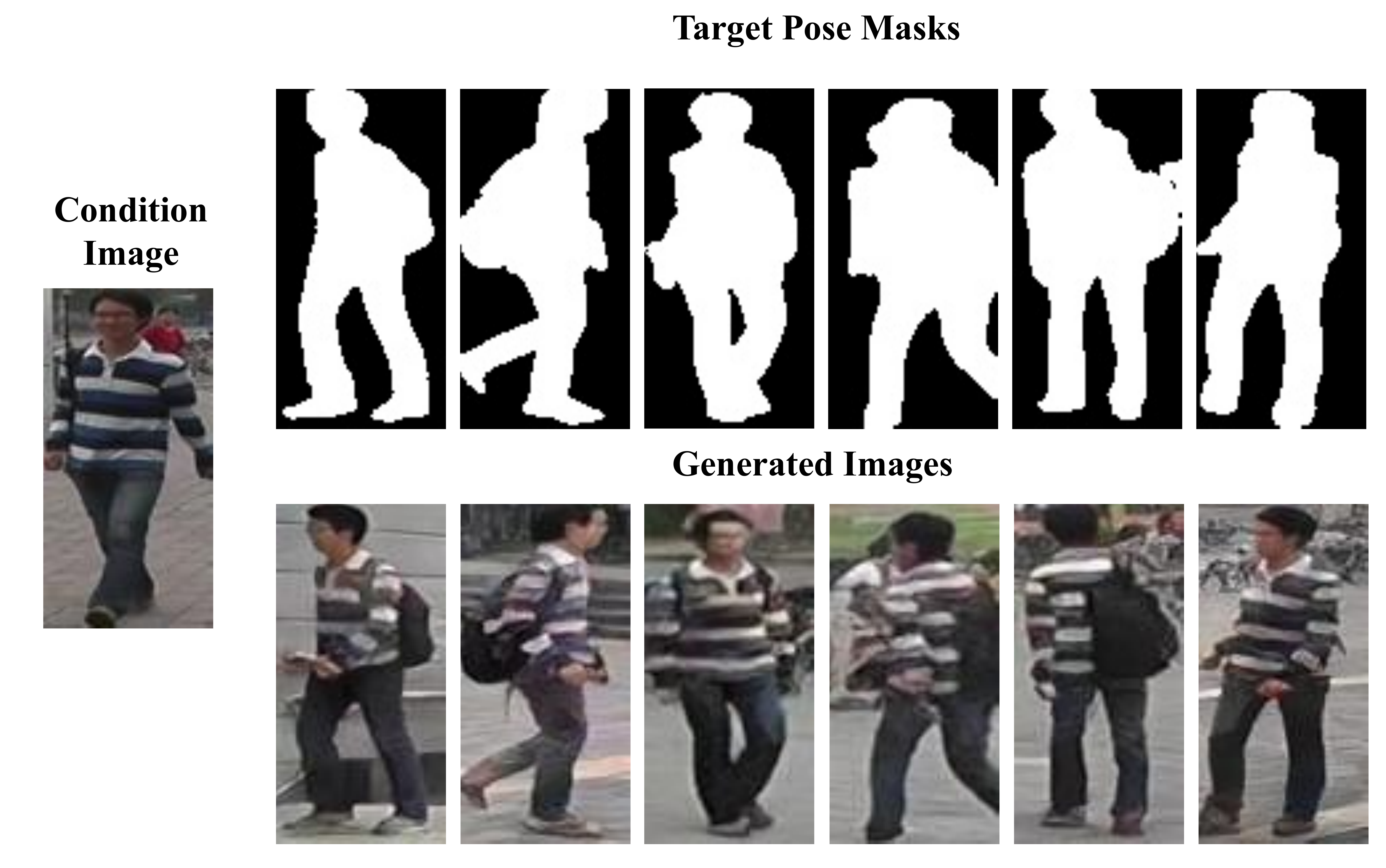}
\caption{Some generated examples by our method based on different target poses.}
\label{fig:1}
\end{figure}

Person re-identification (Re-ID) refers the task of matching a specific person across multiple non-overlapping cameras. It has been receiving considerable attention in the computer vision community due to its various surveillance applications. With the strong learning capabilities of deep neural networks, recent methods \cite{bai2020deep,zheng2019pose,yao2019deep} about person Re-ID has a great process. However, since the existing benchmarks contain a limited number of pose changes, the variations in pose becomes one of the key factors which prevents the network from learning a robust Re-ID model. Motivated by the above discussion, in this paper, we are interested in transferring a person from one pose to another by giving a condition image as show in Fig. \ref{fig:1}. It can provide additional labeled samples for training discriminative methods as a data augmentation tool.

The recent advent of generative adversarial networks \cite{goodfellow2014generative,xing2019unsupervised,xing2020inducing} has provided powerful tools to achieve pose-guided image generation, and inspired many researches in this field by using the generated image to enhance the generalization ability for person Re-ID \cite{siarohin2019appearance,zhu2019progressive,song2019unsupervised,siarohin2019appearance}. However, most of approaches focus on the appearance and the body shape of the person in the pose transfer, where the details of the clothing texture are not fully considered. Due to the low-resolution of the cameras, person Re-ID is usually based on the colors and the clothing attributes to match each other. These clothing attributes are significance for human visual perception.

Ignoring non-rigid human body deformations and clothing shapes can result in compromised quality of the synthesized images. Therefore, how to simultaneously preserve the appearance and clothing attributes of the condition image in the pose transfer becomes much more challenging. The key challenges of the pose transfer include the following two aspects: (1) To preserve clothing attributes is difficult in the generation process, particularly when only given partial observation of the person. (2) Due to the non-rigid nature of human body, it is generally difficult to transform the spatially misaligned body parts for the neural networks.

To address above issue, we propose a \textit{Semantic Attention Network} in this paper. In contrast to the other methods \cite{zhu2019progressive,song2019unsupervised,siarohin2019appearance}, our network is divided into two pathway to capture the appearance feature and the semantic parsing information of the original image. The network takes the representations of the human semantic parsing and the image as inputs. The pose transfer is carried out by a sequence of \textit{Semantic Attention Blocks} (SABs). This scheme allows the generator to simultaneously update the appearance code and the clothing texture before achieving the final output. Moreover, human semantic parsing naturally provides a foreground mask, which is important for seamlessly stitching foreground and background in pose-guided image generation.

Specifically, we design an attention mechanism to focus on the change of the clothing shape in each SAB. This attention mechanism allows the generator to interest in better selecting the image regions, which can preserve or suppress information for transferring. The outputs of several sequences of blocks are the updated appearance code and the semantic code. We can reconstruct the pose-transferred image via the similar U-Net \cite{ronneberger2015u}. Moreover, compared with the keypoint-based pose representation, we adopt the binary segmentation mask as the pose representation as show in Fig \ref{fig:1}. On the hand, the pose mask has the capability of removing the background clutters in pixel-level. On the other, the binary segmentation mask featuring only one channel results in decreasing the computation load and the designed parameters considerably in the training process. The main contributions of this paper can be summarized as follows:
\begin{itemize}
\item We propose a semantic attention network to address the challenging the task of the pose transfer. The proposed network can further improve the qualitative result of the generated image by using the semantic attention block (SAB). These blocks can effectively utilize the semantic parsing and the image feature to smo-othly guide the pose transfer process.

\item We first utilize the segmentation mask as the pose representation in the pose transfer. In contrast to the keyp-oint-based pose representation, the segmentation mask has the capability to remove the background clutters in pixel-level, which can make the generator focus on the human body region and decrease the computation burden drastically.

\item Our method has superior performances both in body shape and keeping clothing attributes on challenging benchmarks. The generated images are utilized to enrich the pose variation of the person, and substantially augment person dataset for the person re-identification application.
\end{itemize}





\section{Related Work}
\label{sec:2}
With the development of the generative adversarial networks (GANs) \cite{mirza2014conditional}, image generation can be potentially applied in many fields such as image restoration \cite{pan2020physics,song2020pet}, image retrieval \cite{song2020unified} and cross-domain image generation \cite{li2020semi}. To address the pose variation and improve the generalization ability of ReID models, most of the existing methods focus on the pose-guided image generation. 

The early attempt on pose transfer was achieved in \cite{ma2017pose}, which is composed of two-stage networks. Stage-$\text{I}$ aims to coarsely generate image under the target pose, and Stage-$\text{II}$ considers more appearance details. To further improve the existing work, Ma $et.al$ \cite{ma2018disentangled} disentangled and encoded three modes (the foreground, the background and the pose) into the embedding features, and then decoded them back to the image. However, the spatial deformation between the source and the target is not fully considered in the aforementioned literature, which makes it difficult  to handle the situations with the misaligned appearance. Therefore, recent works \cite{balakrishnan2018synthesizing,siarohin2019appearance,zhu2019progressive} attempted to transform the pixels of the original image to align with the generated image under the target pose. Balakrishnan $et.al$ \cite{balakrishnan2018synthesizing} proposed a method to separate the source image into different parts and reconstruct them by the target pose. Siarohin $et.al$ \cite{siarohin2019appearance} introduced deformable skips connections that require extensive affine transformation computation. Zhen $et.al$ \cite{zhu2019progressive} presented the pose transfer network to infers the regions of interest based on the human pose. On the basis of the above proposed methods, the appearances of the generated images are less realistic-looking, since the output results are from highly compressed features. Instead, we utilize the semantic parsing map to guide the image synthesis with the higher-level structure information.

Inspire by the text-to-image translation methods in \cite{hong2018inferring,johnson2018image}, Dong $et.al$ \cite{dong2018soft} proposed a network, called Soft-Gated Warping-GAN, which first predicted the target segmentation map, and then estimated the transformations by the geometric matcher for rendering the textures. Han $et.al$ \cite{han2019clothflow} utilized a two-stream architecture to extract the appearance flow between the source and the target clothing regions for the person image generation. Similarly, Song $et.al$ \cite{song2019unsupervised} decomposed the conversion process into two stages: the semantic parsing transformation and the appearance generation. It is worth noting that the predicted failure semantic parsing map in the above literature can affect the quality of the synthesized image.

Compared with the existing work, we introduce a semantic map with an attention mechanism to preserve texture synthesis between the corresponding semantic map areas in supervision manner. And we can generate the person appearances and the clothing texture simultaneously by using the proposed two-pathway network. Furthermore, most of these methods employ the pose and key-points estimation, whereas the binary segmentation mask \cite{song2018mask} based on the pose representation is much cheaper and more flexible. Therefore, we utilize the binary segmentation mask as the target pose and remove the background clutter of the original image.

\section{The Proposed Approach}
\label{sec:3}
\begin{figure*}
\centering
\includegraphics[scale=0.15]{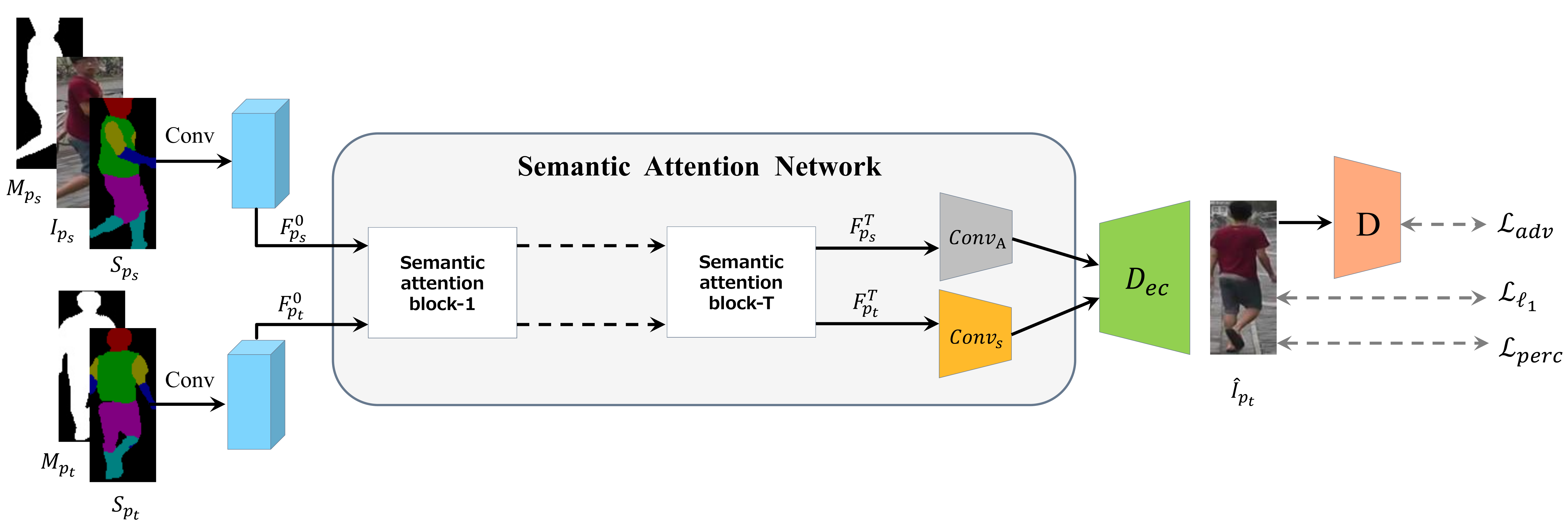}
\caption{Our overall framework for the proposed method. We forward initial inputs into the convolutional layers to obtain the encoding feature $F^0_{p_s}$ and $F^0_{p_t}$, then updated them by the semantic attention network (in gray). The decoder (in green) generates the fake image $\hat{I}_{p_t}$. We utilize the discriminator (in pink) to judge how likely $\hat{I}_{p_t}$ contains the same person in $I_{p_s}$. The detailed illustration of the semantic attention block can be found in Fig. \ref{fig:3}}.
\label{fig:2}
\end{figure*}

\begin{figure}
\centering
\includegraphics[scale=0.15]{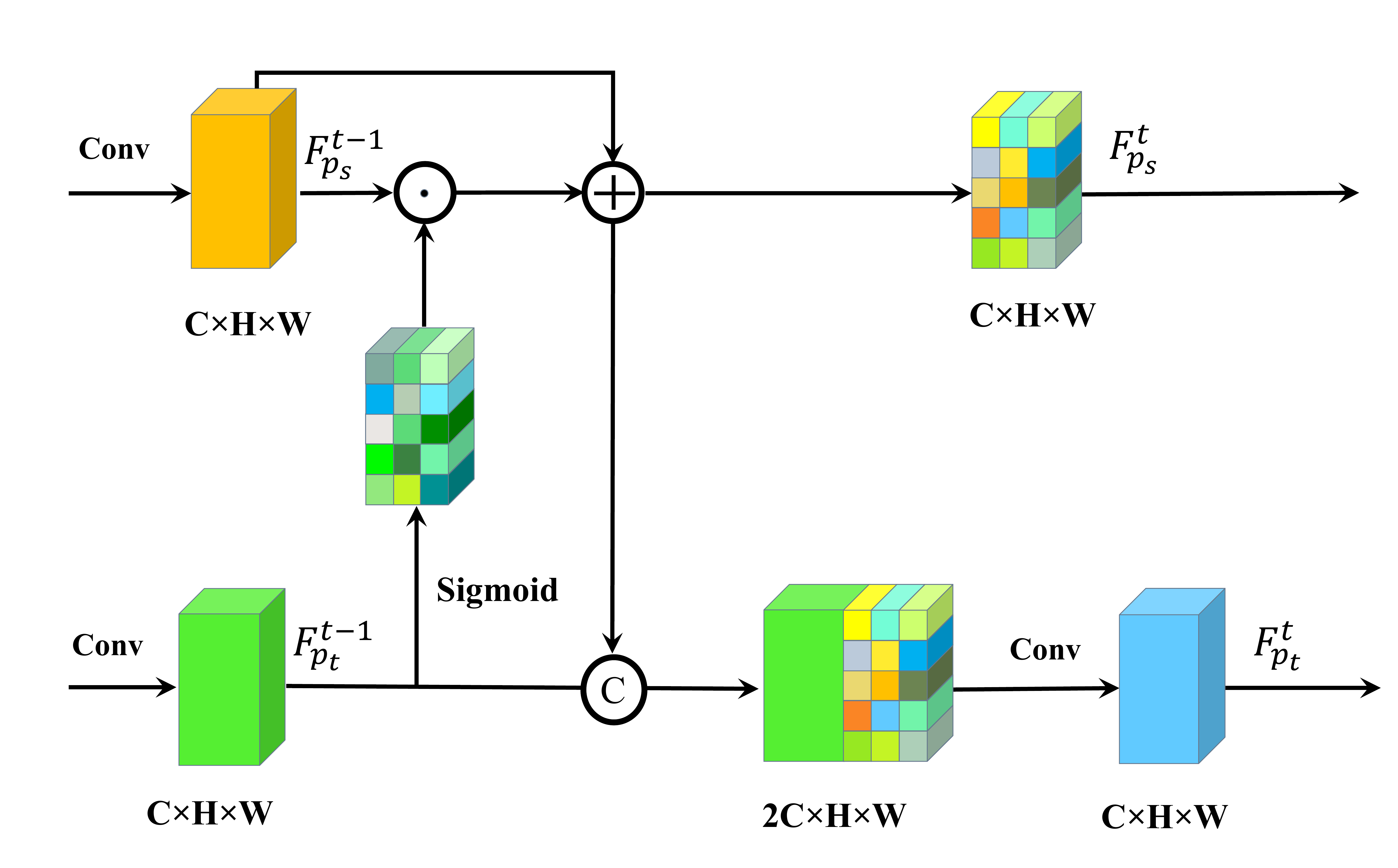}
\caption{The details of the Semantic Attention Block. \textcircled{\textperiodcentered}, \textcircled{+} represent the element-wise multiply and element-wise plus, respectively. \textcircled{c} denotes the depth concatenation.}
\label{fig:3}
\end{figure}
{\bfseries Notations.} We first define the notations to be used in this paper. Assume we have the access to a set of $N$ images $I=\{I_i\}_{i=1}^N$, where $I_i \in \mathbb{R}^{3\times H \times W}$ represents the $i^{th}$ image in the dataset. $S_i$ and $P_i$ are the corresponding semantic map and binary segmentation mask of $I_i$, respectively. The semantic map $S_i$ is extracted by the Pascal-Person-Part \cite{li2019self}. We describe $S_i$ by using a pixel-level one-hot encoding, $i.e.$, $S_i \in \{0,1\}^{L \times H \times W}$, where $L$ denotes the total number of semantic labels. The binary segmentation mask $P_i$ is obtained by Mask-RCNN \cite{he2017mask}, where $P_i \in \mathbb{R}^{H \times W}$ as shown in Fig. \ref{fig:1}.

{\bfseries Problem Formulation.} In this paper, we propose the \textit{semantic attention network} to address the challenging task of the pose transfer. The overview of the proposed method is shown in Fig. \ref{fig:2}. Given a reference image $I_{p_s}$ under the source pose $p_s$, our method aims to transfer the pose of the person image $I_{p_s}$ from the source pose $p_s$ to the target pose $p_t$. The generated image $\hat{I}_{p_t}$ follows the clothing appearance of $I_{p_s}$ but under the pose $p_t$. More details of each component of the proposed network can be found in the following section.
\subsection{Generator}
\label{sec:3.1} 
\subsubsection{Input}
\label{sec:3.1.1}
The generator aims to synthesize the appearance and texture of the output image $\hat{I}_{p_t}$ under the target pose $p_t$, guided by the reference image $I_{p_s}$ and the target semantic map $S_{p_t}$. Therefore, the initial input can be divided into two pathways for encoding features, which are called appearance pathway and semantic pathway, respectively. The appearance pathway is used to extract the appearance details, which takes the reference image $I_{p_s}$, the pose mask $M_{p_s}$ and the semantic map $S_{p_s}$ as input. The another pathway is employed to update clothing textures, which takes $M_{p_t}$ and $S_{p_t}$ as inputs. The inputs of each pathway are stacked along their depth axes before being encoded by a $7\times7$ and two $3\times3$ convolutional layers, each followed by a normalization layers (BN) \cite{ioffe2015batch} and a leaky rectified linear unit (LeakyReLU)\cite{xu2015empirical}. This encoding process mixes the corresponding poses and semantic maps, which features the ability of preserving their pixel-level information and capturing their dependencies. 
\subsubsection{Semantic Attention Network }
\label{sec:3.1.2}
To concentrate on the human body shape and clothing texture transfer, we propose a semantic attention network (SAN), which consists of several semantic attention blocks to infer the interest regions based on the target pose and the appearance. The specific details of the block are shown in Fig. \ref{fig:2}. The encoding features $F^0_{p_s}$ and $F^0_{p_t}$ are simultaneously sent to the SAN. Then, SAN updates these code to obtain the final appearance code $F^T_{p_s}$ and semantic code $F^T_{p_t}$ through the sequence of blocks. We describe the detailed update process as follows.

{\bfseries Appearance Code Update.} The pose transfer is about moving body patches from the reference location to the target location, which can cause a large variations of the articulation and the clothing texture. To address this issue, the attention mask $M$ is used to evaluate the importance of each position. The encoding feature $F^0_{p_t}$ firstly goes through three convolutional layers with two normalization layers and LeakyReLU, then mapping the semantic code $F^{t-1}_{p_t}$ to the values between 0 and 1 by an element-wise sigmoid function:
\begin{equation}
M=\frac{1}{1+e^{x^z_{ij}}},
\end{equation}
where $x^z_{ij}$ denotes the value of the $(i,j)$ element
in $z^{th}$ channel of the updated semantic code. The appearance code $F^t_{p_s}$ are either preserved or suppressed information by multiplying the semantic code $F^{t-1}_{p_t}$ with the attention mask $M$. In addition, we adopt the residual connection to preserve the original image information in the pose transfer. The transformed image feature is added after the element-wise product. The process can be described as:
\begin{equation}
F^t_{p_s} = M_t \odot F^{t-1}_{p_t} + F^{t-1}_{p_s},
\end{equation}
where $M_t$ denotes the attention score of the $t^{th}$ semantic attention block, and $\odot$ denotes the element-wise product. 

{\bfseries Semantic Code Update.} In the pose transfer, the semantic code $F^{t-1}_{p_t}$ need to be updated by incorporating with the new appearance code. The updated method can evaluate which clothing patches can be preserved and how to change the texture in the new poses. Specifically, 
the encoding feature $F^{0}_{p_t}$ is first fed into three convolutional layers with two normalization layers  and LeakyReLU to obtain the semantic code $F^{t}_{p_t}$. Then, $F^{t}_{p_t}$ is concatenated with the updated new appearance code $F^t_{p_s}$ along the depth axis. 
The update process can be expressed as:
\begin{equation}
F^{t}_{p_t} = Conv(F^{t-1}_{p_t}||F^{t}_{p_s}),
\end{equation}
where $||$ denotes the concatenation of the update code. We forward the final updated code $F^{T}_{p_S}$ and $F^{T}_{p_t}$ to $Conv_A$ and $Conv_S$ to obtain their content features, simultaneously. The $Conv_A$ has three down-sampling convolution subnetworks, where each subnetwork consists of the convolutional layers with LeakyReLU and BatchNorm. The architecture of $Conv_S$ is similar to $Conv_A$.

\begin{figure}
\centering
\includegraphics[scale=0.15]{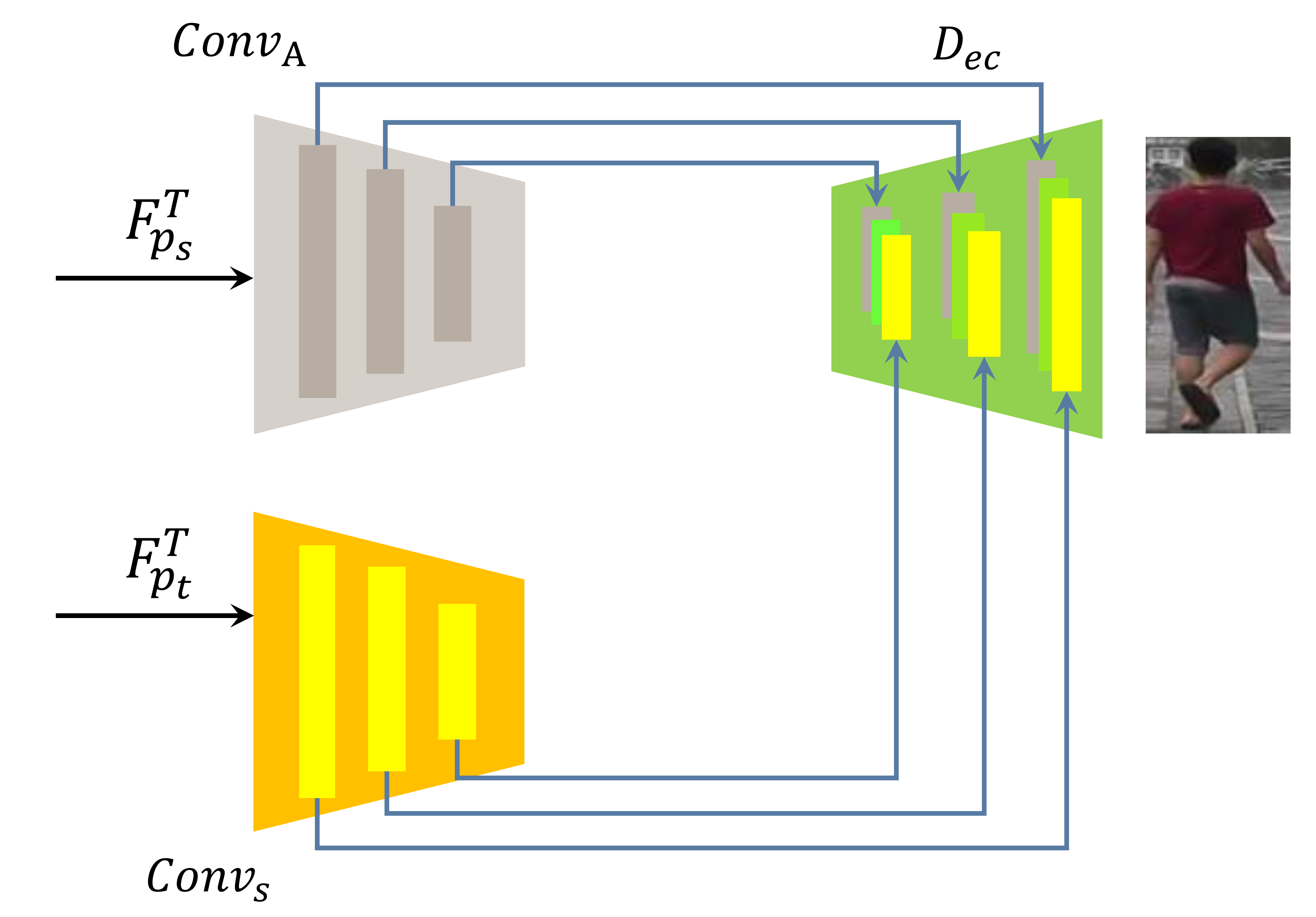}
\caption{The details of the Semantic Attention Block. \textcircled{\textperiodcentered}, \textcircled{+} represent the element-wise multiply and element-wise plus, respectively. \textcircled{c} denotes the depth concatenation.}
\label{fig:8}
\end{figure}
\subsubsection{Decoder}
\label{sec:3.1.3}
The structure of the decoder is similar to U-net \cite{ronneberger2015u}, which can generate the image $\hat{I}_{p_t}$ under the target pose $p_t$ based on the encoded feature.  At each up-sampling step, semantic features of $Conv_S$ can be concatenated with the same size feature map in the corresponding channel of $Conv_A$, which can learn how to assemble the more realistic-looking output image based on this information by $N$ deconvolutional layers.

\subsection{Discriminator}
\label{sec:3.3}
We employ a discriminator $D$ to help generator network to differentiate the real and fake image. The discriminator $D$ takes $\hat{I}_{p_t}$ and $I_{p_t}$ as inputs, which is built by three residual blocks after two down-sampling convolutions. The experiments have shown that if we use the hard label ($i.e.$ 0 or 1), the loss value of the discriminator will quickly fall to zero. Accordingly, a sigmoid layer is employed at the end of the discriminator to obtain the probabilities output.

\subsection{Loss Function}
We train the network using a joint loss consisting of a adversarial loss, reconstruction $L_1$ loss, and perceptual loss.

{\bfseries Adversarial Loss.} The generative adversarial framework is employed to simulate the distributions of the ground-truth $I_{p_t}$. The generator loss can be written as:
\begin{align}
\mathcal{L}_{adv}=&{\mathbb{E}_{({I_{p_s},{I_{p_t}})}\in{\mathcal{P}}}}[\log D(I_{p_s},I_{p_t})]\notag\\
&+{\mathbb{E}_{{{I_{p_t}}\in{\mathcal{P}},{\hat{I}_{p_t}}\in{\hat{\mathcal{P}}}}}}[\log(1-D({I_{p_t}},\hat{I}_{p_t}))].
\end{align}
Note that ${\mathcal{P}}$ and ${\hat{\mathcal{P}}}$ denote the distribution of the ground-truth images and the synthesized images, respectively. ${I_{p_t}}$ represents the real target image under the pose $p_t$.

{\bfseries $\ell_1$ Loss.} The $\mathcal{L}_{\ell_1}$ loss can be further written as:
\begin{equation}
\mathcal{L}_{\ell_1}=\|{I}_{p_t}-\hat{I}_{p_t}\|_1,
\end{equation}
where $\mathcal{L}_{\ell_1}$ denotes the pixel-wise $L_1$ loss computed between the synthesized image $\hat{I}_{p_t}$ and the target ground-truth image $I_{p_t}$. 

{\bfseries Perceptual Loss.} To generate more smooth and  realistic-looking person image, we also utilize the perceptual loss, which is the Euclidean distance between feature representations:
\begin{equation}
\hspace{-0.3cm}\mathcal{L}_{perc}\!=\!\frac{1}{{W_\rho}{H_\rho}{C_\rho}}\!\sum^{W_\rho}_{x=1}\!\sum^{H_\rho}_{y=1}\!\sum^{C_\rho}_{z=1}\!\|\phi_\rho(\hat{I}_{p_t})_{x,y,z}\!-\phi_\rho(I_{p_t})_{x,y,z}\|^2_2,
\end{equation}
where $\phi_\rho$ is the activation map of the $\rho^{th}$ layer of the pre-trained VGG-19 model \cite{simonyan2014very} on ImageNet \cite{russakovsky2015imagenet}. $W_\rho$, $H_\rho$ and $C_\rho$ denote the spatial width, height and depth of $\phi_\rho$, respectively.

The overall loss function of the model as:
\begin{equation}
\mathcal{L}_{full}=\arg\mathop{\min}\limits_{G}\mathop{\max}\limits_{D}\alpha\mathcal{L}_{adv}+\beta \mathcal{L}_{\ell_1}+\gamma\mathcal{L}_{perc},
\end{equation}
where $\mathcal{L}_{adv}$ denotes the adversarial loss, $\mathcal{L}_{L_1}$ denotes the pixel-wise $L_1$ loss and $\mathcal{L}_{perc}$ is the perceptual $L_1$ loss. $\lambda$, $\beta$ and $\gamma$ represent the weight factors of three loss terms that contributes to $\mathcal{L}_{full}$, respectively.


\section{Experiments}
In this section, we conduct extensive experiments to verify its design rationalities and efficiency of the proposed network. The experiments demonstrate the superiority of our method in both objective quantitative scores and subjective visual realness.
\subsection{Datasets and Metrics}
\label{sec:4.1}
{\bfseries Datasets.}
We implement experiments on the challenging person Re-ID dataset Market-1501 \cite{zheng2015scalable} and the \textit{In-shop Clothes Retrieval Benchmark} of the DeepFashion dataset \cite{liu2016deepfashion}. Market-1501 contains 32,668 low-resolution images of $128\times64$, which has vary enormously in the pose, viewpoint, background and illumination. DeepFashion contains 52,712 high-resolution images of $256\times256$, which has contains a large number of clothing images with the various pose and the appearance. We randomly collect 243,200 training pairs and 12,000 testing pairs for Market-1501, and collect 106,966 training pairs and 8,570 testing pairs for DeepFashion. The person identities of the training sets and the testing sets do not overlap, which better evaluate the generalization ability of the network.

{\bfseries Metrics.}
In our experiments, we adopt the Learned Perceptual Image Patch Similarity (LPIPS) \cite{zhang2018unreasonable} and Fr´echet Inception Distance (FID) \cite{heusel2017gans} for the quantitative evaluation. LPIPS calculates the reconstruction error between the synthesized images and the ground-truth images at the perceptual domain. We also use its masked version, mask-LPIPS, to reduce the background influence by masking it out. FID computes the Wasserstein-2 distance between distributions of the synthesized images and the reference images. 

\subsection{Implementation details}
\label{sec:4.2}
Our method is implemented by using the popular Pytorch framework. For the person representation, the semantic maps are extracted by the Pascal-Person-Part \cite{li2019self}. We employ the semantic label originally defined in \cite{li2019self} and set $L = 20$ ($i.e.$ background, hat, hair, glove, sunglasses, upper-clothes, dress, coat, socks, pants, jumpsuits, scarf, skirt, face, left-arm, right-arm, left-leg, right-leg, left-shoe, right-shoe.) We utilize 5 semantic blocks in the network for both datasets. For Market-1501, we directly employ the training set and the testing set on $128\times64$. The proposed method is trained by using the Adam optimizer \cite{kingma2014adam} with $\beta_1 = 0.5$, $\beta_2 = 0.999$. Learning rate is initially set to $2\times10^{-4}$, and linearly decay to 0 after 300 epochs. The batch size is set to 32. The $\alpha$, $\beta$, and $\gamma$ are set to 10, 15, and 5. For DeepFashion, we utilize the training set and the testing set on $256\times256$. Learning rate is initially set to $2\times10^{-4}$, and linearly decay to 0 after 500 epochs. The batch size is set to 8. The $\alpha$, $\beta$, and $\gamma$ are set to 15, 1, and 5.

\begin{figure}
\centering
\includegraphics[width=0.49\textwidth]{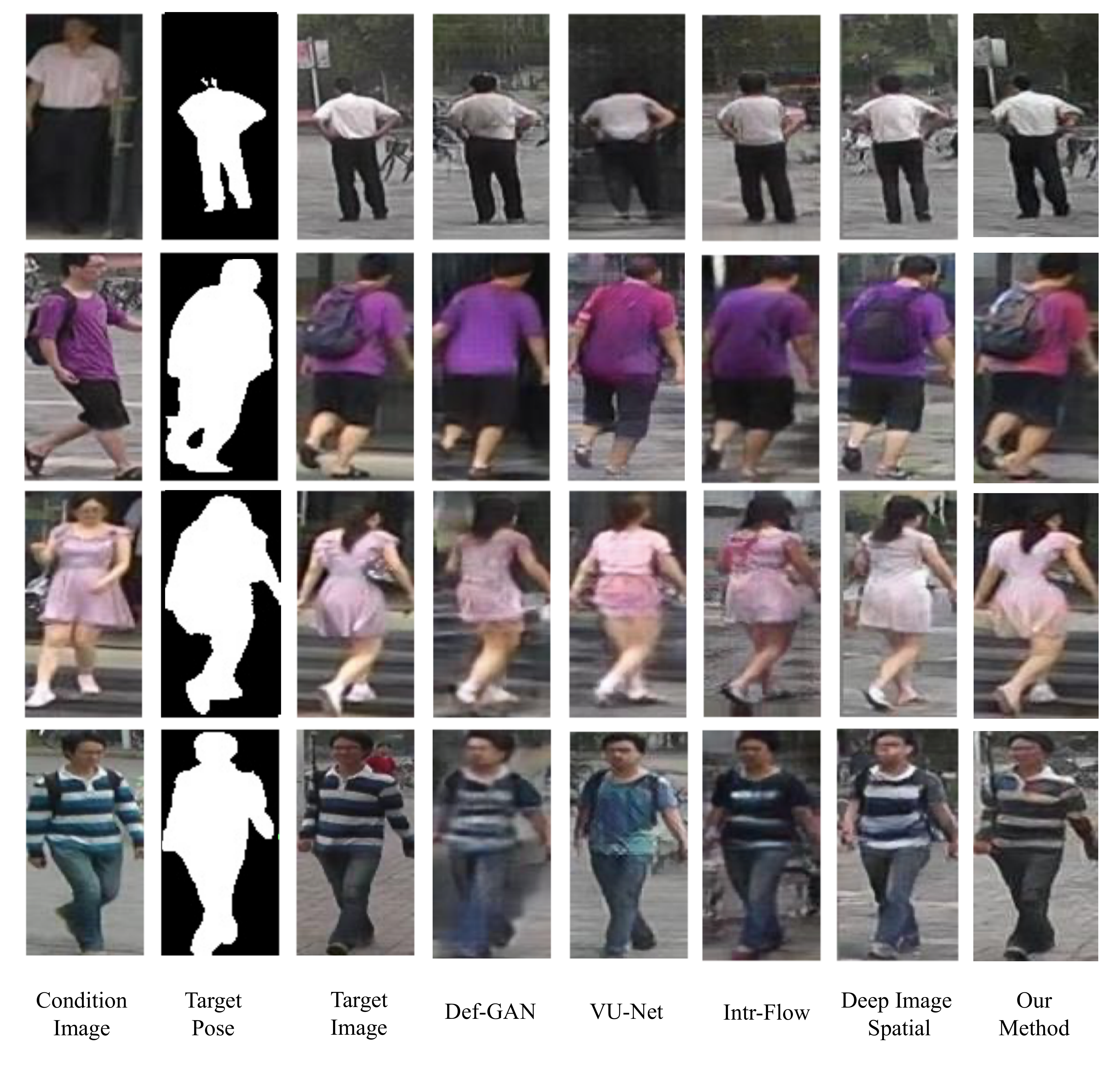}
\caption{Compared with different methods on Market-1501 dataset, which are Def-GAN \cite{siarohin2019appearance}, VU-Net \cite{esser2018variational}, Intro-Flow \cite{li2019dense}, Intro-Flow \cite{li2019dense}, Deep Image Spatial \cite{ren2020deep}.}
\label{fig:4}
\end{figure}

\begin{figure}
\centering
\includegraphics[width=0.49\textwidth]{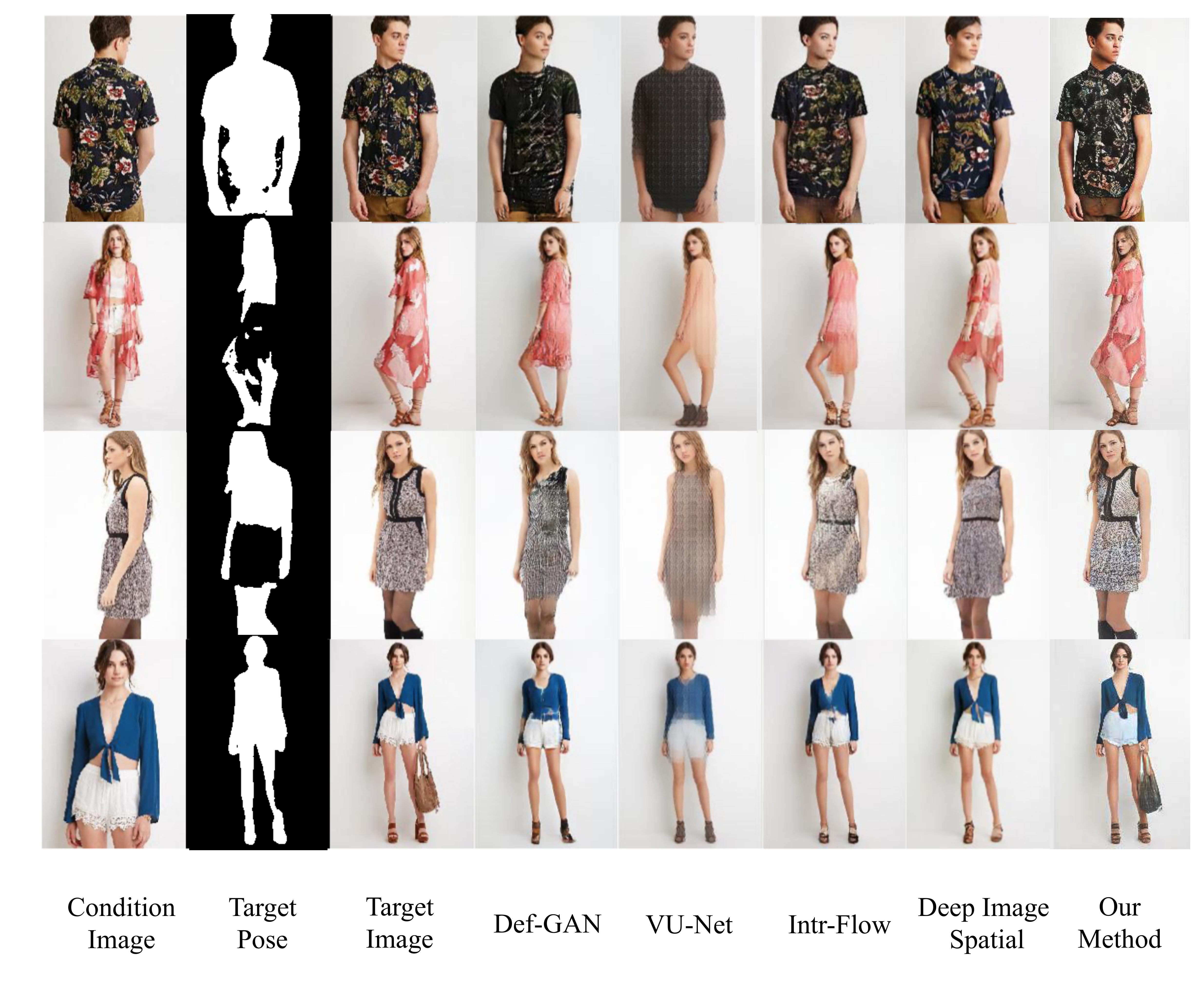}
\caption{Compared with different methods on DeepFashion dataset, which are Def-GAN \cite{siarohin2019appearance}, VU-Net \cite{esser2018variational}, Intro-Flow \cite{li2019dense}, Intro-Flow \cite{li2019dense}, Deep Image Spatial \cite{ren2020deep}.}
\label{fig:5}
\end{figure}
\subsection{Comparison with previous work}
\begin{table*}
\centering
\caption{Comparison with quantitative results on Market-1501 and DeepFashion datasets.}
\label{tab:1}       
\begin{tabular}{c||c c c||c c}
\hline
\multirow{2}{*}{Methods} & \multicolumn{3}{c||}{Market-1501} & \multicolumn{2}{c}{DeepFashion} \\  \cline{2-6}
& FID            & LPIPS          & Mask-LPIPS              &     FID       & LPIPS       \\ \hline 
VU-Net \cite{esser2018variational}              & 20.144        & 0.3211         & 0.1747           & 23.667           & 0.2637  \\ \hline
Def-GAN \cite{siarohin2019appearance}              & 25.364        & 0.2994         & 0.1496           & 18.457     & 0.2330 \\ \hline
Intro-Flow \cite{li2019dense}              & 27.163        & 0.2888         & 0.1403           & 16.314           & 0.2131  \\ \hline
Deep Image Spatial \cite{ren2020deep}              & 19.751         & 0.2817         & 0.1482           & 10.573           & 0.2341  \\ \hline
Our method               & \textbf{14.142}         &          \textbf{0.2632}  &\textbf{0.1282}           &   \textbf{8.732 }        &    \textbf{0.2078}   \\ \hline
\end{tabular}
\end{table*}
\label{sec:4.3}
{\bfseries Quantitative Result.} We compare our model  with several stare-of-the-art methods including VU-Net \cite{esser2018variational}, Def-GAN \cite{siarohin2019appearance}, Intro-Flow \cite{li2019dense} and Deep Image Spatial \cite{ren2020deep}. The quantitative comparison results are shown in Tab.\ref{tab:1}. To the fair comparison, we download the pre-trained models of the previous works and re-evaluate the quantitative performances on our testing set. In contrast to the other methods, our proposed method achieves the best results in both datasets, which means that our model can generate realistic results with fewer reconstruction errors. The generated images in our model have more realistic details and better body shape.

{\bfseries Qualitative Comparison.} We evaluate the performance of our network both on Market-1501 and DeepFashion dataset. Some typical qualitative examples on Market-1501 are shown in Fig. \ref{fig:4}. Our method achieve superior in the appearance and the clothing attributes. For Market-1501, especially, the cloth color in the first row, the detailed bag in the second row, and the clearly clothing textures in the last two row. For DeepFashion, the clear collar edge of the T-shirt in the first row, the detailed clothes texture in the second and third row, the completed shoes and bag in the last row.

\subsection{Ablation Study and Qualitative Analysis}
\label{sec:4.4}
\subsubsection{Analysis of each part of our method.}
\label{sec:4.1.1}
In this section, we present an ablation study to evaluate the contribution of each component. The discriminator architecture is the same for all the methods.

{\bfseries PercLoss.} Perceptual $\ell_1$ loss is effective in super resolution, style transfer as well as the pose transfer task. For the evaluate the impotent of the perceptual $\ell_1$ loss, we only train the network using a joint loss consisting of a adversarial loss, reconstruction $\ell_1$ loss.

{\bfseries Segmentation Mask.} The binary segmentation mask to directly separate the background of the input images, which can help the generator and the discriminator focus on the human body region. In order to verify the effective of the binary segmentation mask, we remove the segmentation mask from the initial inputs. All the network as same as full pipeline.

{\bfseries Full pipeline.} We use our proposed the semantic attention framework in this model.
\begin{figure}
\centering
\includegraphics[width=0.49\textwidth]{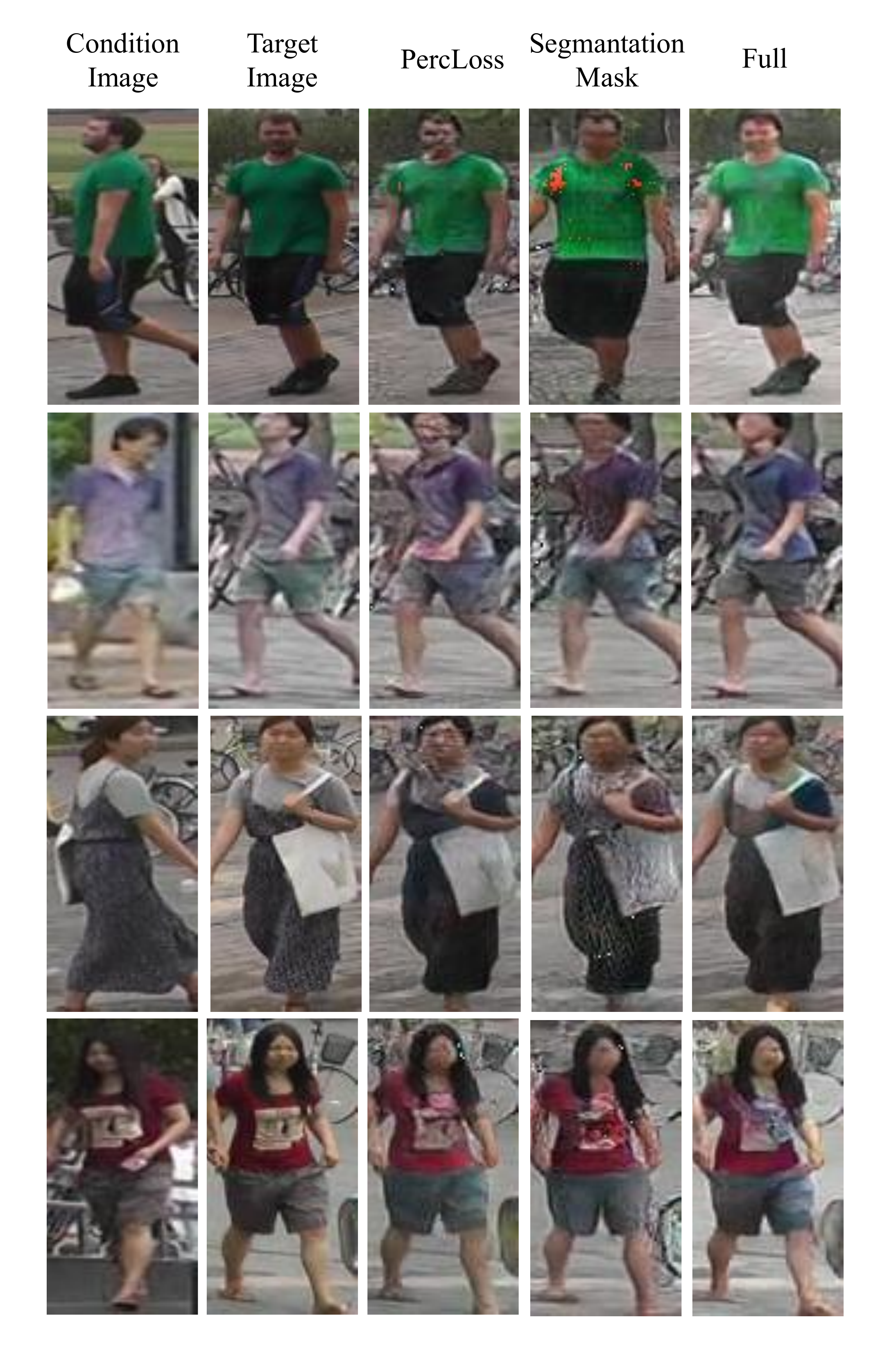}
\caption{Qualitative results on the Market-1501 dataset with respect to different variants of our method.}
\label{fig:7}
\end{figure}

Fig \ref{fig:7} shows the corresponding generated image of the different variants of our method. We analyze the perceptual loss function in our network. In same case, the improvement of Full with respect to PrecLoss is quite drastic, such as drawing on the face. Moreover, without the guidance of the segmentation mask, the clothing texture can be confused with background. The network is difficult to handle the cloth attribute and the body shape at the same. The introduction of the segmentation mask and perceptual loss function can improve the visual quality of the output image.

\subsubsection{Analysis of the semantic attention blocks.}
\label{sec:4.1.2}
\begin{figure}
\centering
\includegraphics[width=0.49\textwidth]{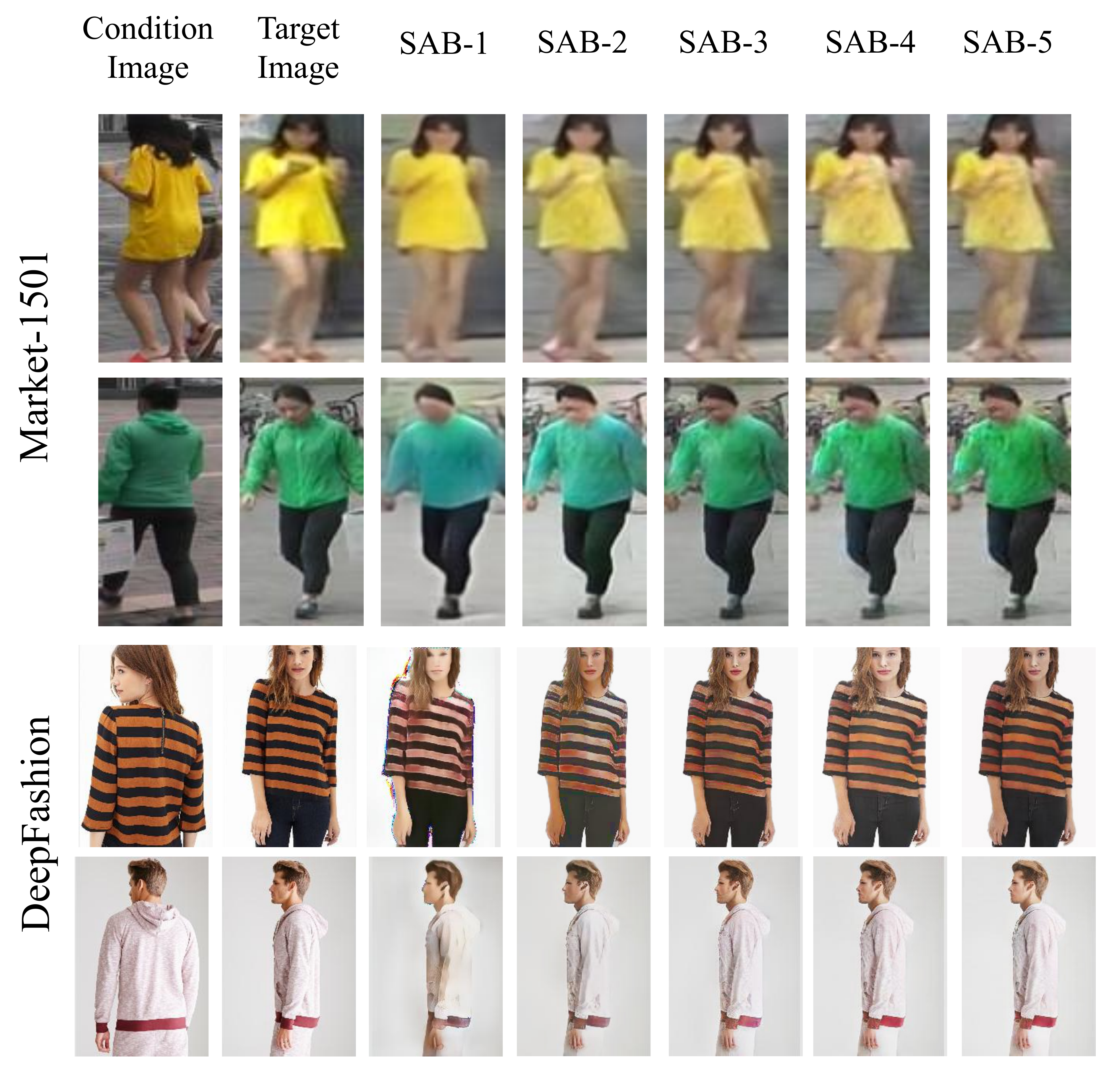}
\caption{Qualitative results of the different numbers of the SAB on Market-1501 and DeepFashion datasets.}
\label{fig:6}
\end{figure}
In this section, we analysis the effective of the semantic attention blocks. We consider the impact of the different numbers of attention blocks on the network.  The semantic attention block aims to capture the appearance consistency and semantic information, simultaneously. We can improve the ability of the generator by varying the number of SABs. Qualitative comparison result are shown in Fig \ref{fig:6}. As the number of SAB increase, the appearances of the person are becoming clear, and the clothing textures are gradually as same as the target image. The overall quality of the image are refined from the foreground to background.

\section{Application to person re-identification}
\label{sec:5}

Person re-identification (Re-ID) aims to match person across no-overlapping video camera. This task has attracted considerable attention for its application in automatic video surveillance. Since the existing benchmarks such as Market-1501 contain a limited number of pose changes, pose variation becomes one of the key factors which prevents the network from learning a robust Re-ID model. The experiments in this section are motivated by the importance of using generative methods as a data-augmentation tool which provides additional labeled samples for training discriminative methods.

In this section, we utilize the different person Re-ID approaches to evaluate the performance of our method in augmenting the person Re-ID on Market-1501. Identification Embedding (IDE) \cite{zheng2016person}) is an approach that regards Re-ID training as an image classification task. In the test phase, the identity of the person is assigned based on the feature representation, where each image feature is obtained by the classification layer of the network. Each query image is associated to the identity of the closest image in the gallery. In our experiment, we employ different metrics to calculate the distance among the feature representations, which are Euclidean distance \cite{gower1985properties}, Cross-view Quadratic Discriminant Analysis (XQDA \cite{gower1985properties}) and a Mahalanobis-based distance (KISSME \cite{koestinger2012large}). The another approach, Siamese network \cite{zheng2017discriminatively} predicts whether the identities of the two input images are the same. For all approaches, we adopt pre-trained ResNet-50 \cite{he2016deep} as test benchmark. 

We augment the Market-1501 dataset by a factor $\alpha$. For each person image, we should select $\alpha-1$ target poses to guide the generate network to synthesize new image. Each synthesized image is labeled with the identity of the reference image. We compare our model performance to several existing person image generators \cite{esser2018variational,ma2018disentangled,siarohin2019appearance,zhu2019progressive} under the same setting for Re-ID data augmentation. Compared with the previous methods, our method can achieve competitive results as shown in Tab. \ref{tab:2}. Our proposed method can generate more smooth and nature human images and be more effective for the Re-ID task.

\section{Conclusion}
In this paper, we propose a semantic attention network for person image generation to deal with the challenging pose transfer. To address the complexity of learning the clothing attributes under different poses, we design a attention mechanism to capture the pose and semantic parsing simultaneously. We take the binary segmentation and the human semantic map as the network input, which can reduce the effect of the background clutter and decrease the computation load. Compared with previous works, our model exhibits the superior performance in both the quantitative scores and the visual realness. Moreover, the experiment results show that our proposed method can be used to alleviate the insufficient training data problem for person Re-ID substantially. In the future, we will predict the pose mask and the semantic map of the desired pose to support the unsupervised person image generation.

\hspace*{\fill} \\

{\bfseries Acknowledgments}
The work is supported by National Natural Science Foundation of China (61573114) and Fundamental Research Funds for the Central Universities (HEU-CF160415).

\bibliographystyle{cas-model2-names}

\bibliography{cas-refs}





\end{document}